\pdfoutput=1

\documentclass[11pt]{article}

\usepackage[review]{acl}

\usepackage{times}
\usepackage{latexsym}

\usepackage[T1]{fontenc}

\usepackage[utf8]{inputenc}

\usepackage{microtype}

\usepackage{hyperref}       
\usepackage{url}            
\usepackage{booktabs}       
\usepackage{amsfonts}       
\usepackage{nicefrac}       
\usepackage{microtype}      
\usepackage{makecell}
\usepackage{graphicx}
\usepackage{subfig}
\usepackage{float}
\usepackage{multirow}
\usepackage{bm}
\usepackage{amsopn}
\usepackage{amsmath}
\usepackage{xcolor}
\usepackage{color}
\usepackage{wrapfig}
\usepackage{multibib}
\usepackage{booktabs}
\usepackage{verbatim}
\usepackage{colortbl}
\usepackage{makecell}

\newcommand{\graycell}{\rowcolor[gray]{.90}}

\title{Input-Tuning: Adapting Unfamiliar Inputs to Frozen Pretrained Models}

\author{
  Shengnan An$^{1}$\thanks{\, Work done during an internship at Microsoft Research. The first two authors contributed equally to this paper.}, Yifei Li$^{2*}$, Zeqi Lin$^{3}$, Qian Liu$^{4}$, Bei Chen$^3$, \\
  \textbf{Qiang Fu$^3$, Weizhu Chen$^5$, Nanning Zheng$^1$, Jian-Guang Lou$^3$}\\
  $^1$ Xi'an Jiaotong University; $^2$ Peking University;\\
  $^3$ Microsoft Research Asia; $^4$ Beihang University; $^5$ Microsoft Azure AI\\
  \small{\texttt{\{an1006634493@stu, nnzheng@mail\}.xjtu.edu.cn}} \\
  \small{\texttt{liyifei@stu.pku.edu.cn}; \quad \texttt{qian.liu@buaa.edu.cn}} \\
  \small{\texttt{\{Zeqi.Lin, beichen, qifu, wzchen, jlou\}@microsoft.com}}
}

\begin{document}
\maketitle

\begin{abstract}

Recently the prompt-tuning paradigm has attracted significant attention.
By only tuning continuous prompts with a frozen pre-trained language model (PLM), prompt-tuning takes a step towards deploying a shared frozen PLM to serve numerous downstream tasks.
Although prompt-tuning shows good performance on certain natural language understanding (NLU) tasks, its effectiveness on natural language generation (NLG) tasks is still under-explored. 
In this paper, we argue that one of the factors hindering the development of prompt-tuning on NLG tasks is the unfamiliar inputs (i.e., inputs are linguistically different from the pretraining corpus).
For example, our preliminary exploration reveals a large performance gap between prompt-tuning and fine-tuning when unfamiliar inputs occur frequently in NLG tasks.
This motivates us to propose \textbf{input-tuning}, which fine-tunes both the continuous prompts and the input representations, leading to a more effective way to adapt unfamiliar inputs to frozen PLMs.
Our proposed input-tuning is conceptually simple and empirically powerful.
Experimental results on seven NLG tasks demonstrate that input-tuning is significantly and consistently better than prompt-tuning.
Furthermore, on three of these tasks, input-tuning can achieve a comparable or even better performance than fine-tuning. 

\end{abstract}

\section{Introduction}

\begin{figure}[t]
    \centering
    \includegraphics[width=0.47\textwidth]{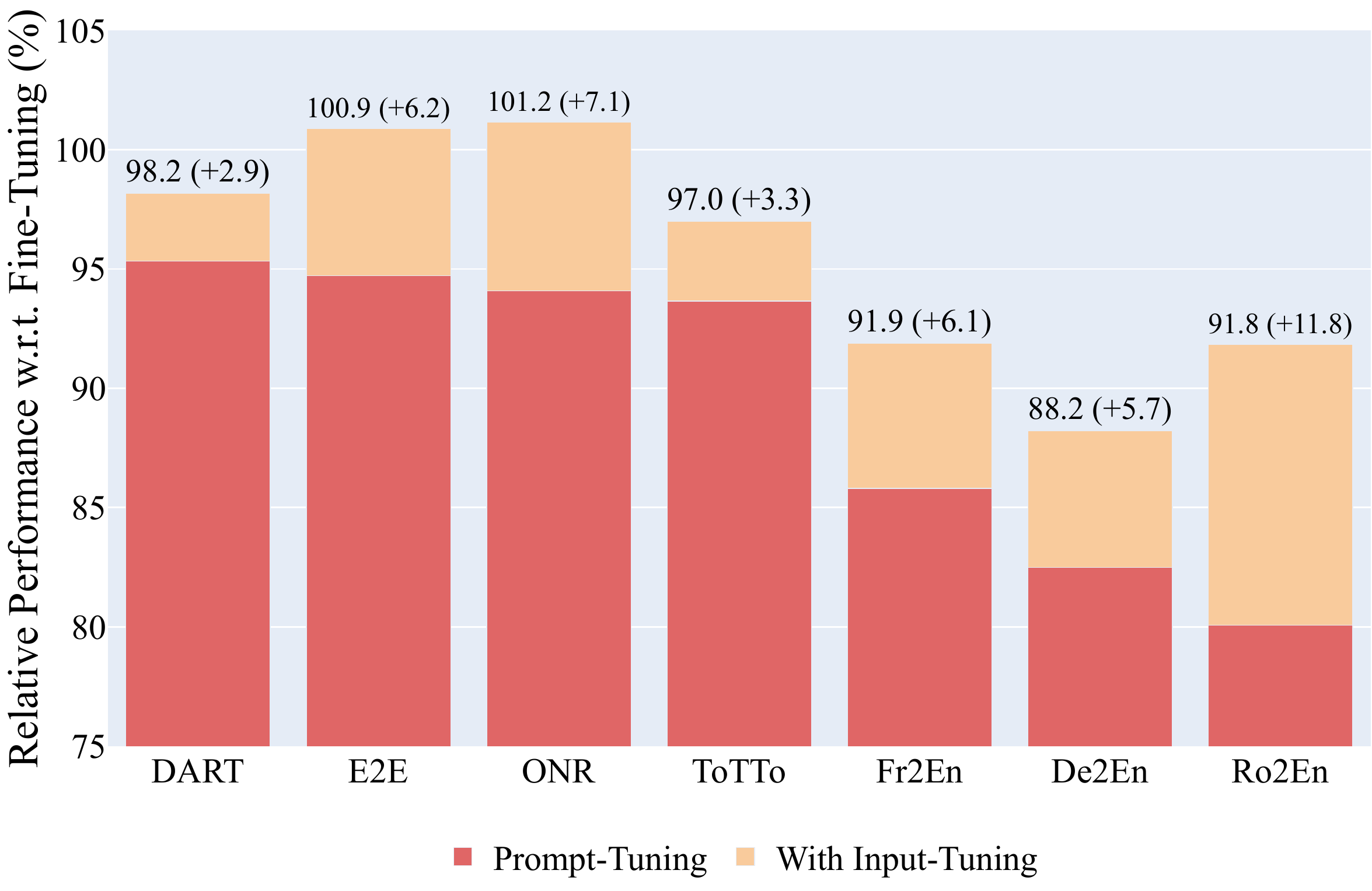}
    \caption{Relative performance (w.r.t. fine-tuning) of prompt-tuning and input-tuning on seven NLG tasks with unfamiliar inputs. The frozen PLM is \textsc{T5-Large} ($\sim$770M parameters) \cite{raffel2020exploring}. The performance of prompt-tuning lags behind that of fine-tuning. Input-tuning (our method) can close this gap significantly and consistently.}
    \label{fig:res}
\end{figure}

\begin{figure*}[thb]
    \centering
    \includegraphics[width=0.75\textwidth]{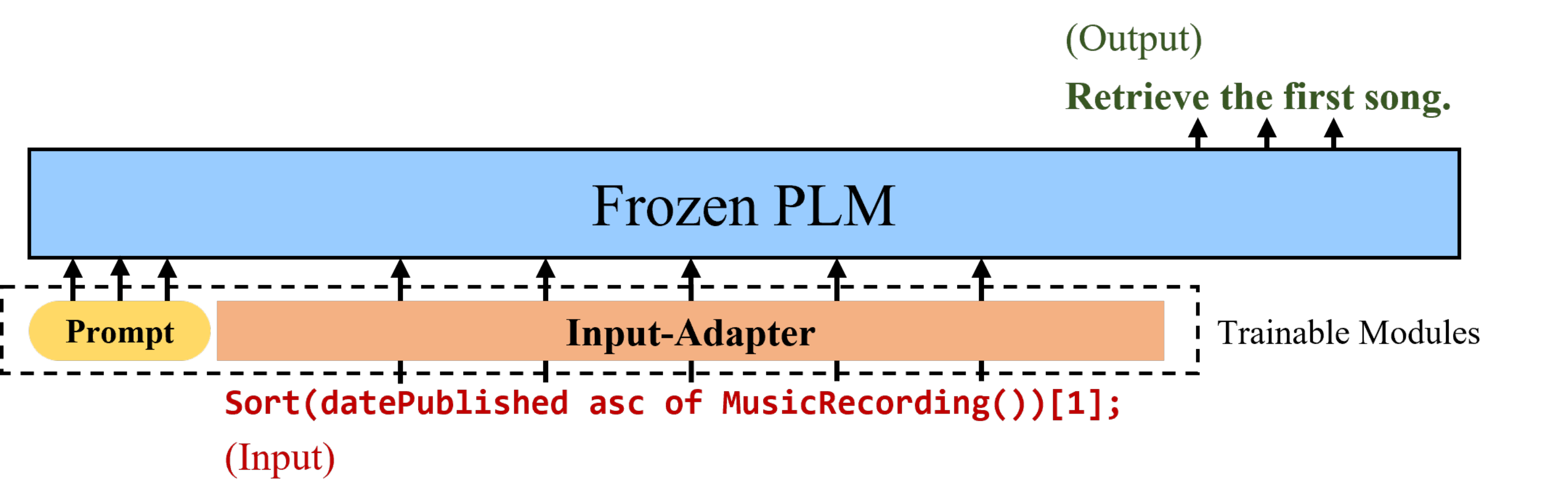}
    \caption{Input-tuning: comparing to prompt-tuning, we propose to add a lightweight trainable module between word embeddings and the bottom layer of the frozen PLM, thus affecting the encoding of unfamiliar inputs more directly and effectively.}
    \label{fig:input_tuning}
\end{figure*}

Recently, there has been a surge of interest in \textbf{prompt-tuning}~\citep{liu2021pre, jiang2020can,shin2020autoprompt,liu2021gpt,lester2021power}.
Unlike fine-tuning that adapts pretrained language models (PLMs) to downstream tasks by tuning all parameters, prompt-tuning is based on the intuition that tasks can be adapted to a frozen PLM via proper task-specific contexts.
Concretely, for each task, prompt-tuning freezes PLM parameters, but optimizes a small continuous task-specific prompt (i.e., a vector sequence) to be prepended to the input text.
Prompt-tuning is appealing, as it takes a step towards deploying a frozen PLM as a common cloud service that can serve many downstream tasks from customers.

The intuition behind prompt-tuning is to cast each downstream task into a language model format that fits the pretraining objective.
This leads to the remarkable success of prompt-tuning on natural language understanding (NLU) tasks~\citep{liu2021gpt,lester2021power}.
However, many tasks, especially natural language generation (NLG) tasks, involve \textbf{unfamiliar inputs}, i.e., \emph{the input sequences are linguistically different from the pretraining corpus}.
For example, in a logic-to-text task, the input sequences are written in some domain-specific logic expressions (e.g. ``\texttt{\small Sort(datePublished asc of MusicRecording())[1];}'') and the output sequences are their natural descriptions (e.g., ``Retrieve the first song'').
Moreover, for PLMs pretrained on English corpus, machine translation tasks with non-English inputs (e.g., French, German, and Romanian) can also be regarded as tasks with unfamiliar inputs.
We can hardly cast such inputs to a format that fits the pretraining objective by simply prompting.
Therefore, it is natural to ask \emph{whether prompt-tuning can be useful at such tasks}, but this has not yet been discussed in the prior work.

Our preliminary exploration shows that, on NLG tasks with unfamiliar inputs, the performance of prompt-tuning lags far behind that of fine-tuning (Figure \ref{fig:res}, detailed in Section \ref{section:experiments}).
Though \citet{lester2021power} reported that, on NLU tasks, prompt-tuning can match the performance of fine-tuning by using giant PLMs (\textgreater{10} billion parameters), our preliminary results indicate that this success can hardly be replicated on tasks with unfamiliar inputs:
for example, on the E2E task, changing the frozen PLM from T5-Large($\sim$770M parameters) to T5-$11$B only improves the BLEU score from $64.5$ to $65.6$, while fine-tuning on T5-Large achieves $68.5$.

In this paper, we first conduct preliminary exploration to empirically show \emph{how the performance of prompt-tuning is limited by unfamiliar inputs}.
In particular, we find that: the gap between prompt-tuning and fine-tuning can be controlled by manually transforming the inputs towards/away the pretraining distribution.
Therefore, to adapt unfamiliar inputs to the frozen PLM effectively, not only prompting, but also transforming their surface representations directly, is required.

This finding indicates the limitation of prompt-tuning and the direction to alleviate it, but it still remains challenging to effectively generalize prompt-tuning to tasks with unfamiliar inputs, since it is not easy to find the best way that the unfamiliar inputs should be transformed.
To mitigate this, we propose \textbf{input-tuning}, which bridges the gap between prompt-tuning and fine-tuning (Figure~\ref{fig:res}) by making the transformation learnable.
Concretely, we \textbf{add a lightweight trainable module between word embeddings and the bottom layer of the PLM, to adjust the encoding of unfamiliar inputs directly} (i.e., the ``input-adapter'' module in Figure~\ref{fig:input_tuning}).
For each task, the soft prompt and the input-adapter are the only two trainable modules, and they are optimized jointly.
During inference, each input sequence is encoded by the input-adapter, concatenated to the soft prompt, and then sent to the frozen PLM to generate an output sequence.

Experimental results on seven NLG tasks demonstrate that input-tuning is significantly and consistently better than prompt-tuning.
Moreover, on three of these tasks, input-tuning can achieve comparable or even better performance than fine-tuning.
We further explore the effectiveness of input-tuning with different backbones, showing that it is applicable with both the encoder-decoder architecture and the auto-regressive language model (Section~\ref{sec:diff_backbones}).
For different data scales, input-tuning stably outperforms prompt-tuning and prefers low-resource scenarios (Section~\ref{sec:low_resource}).

\section{Background}

\subsection{Sequence-to-Sequence Learning}

Sequence-to-sequence learning aims to model the conditional probability of the target sequence $\mathbf{y} = \{y_1, ..., y_m\}$ given the source sequence $\mathbf{x} = \{x_1, ..., x_n\}$:
\begin{equation}
P(\mathbf{y}|\mathbf{x})=\prod_{i=1}^{m}{P(y_i|\mathbf{x}, \mathbf{y}_{< i})},
\end{equation}
where the factor $P(y_i|\mathbf{x}, \mathbf{y}_{< i})$ can be estimated by a neural model $p_\phi(y_i|\mathbf{x}, \mathbf{y}_{< i})$ parametrized by $\phi$.
Suppose that $p_\phi$ is a pretrained language model, either autoregressive language model (such as GPT-2 \cite{brown2020language} and UniLM \cite{dongunified}) or encoder-decoder model (such as T5 \cite{raffel2020exploring} and BART \cite{lewisetal2020bart}).

In the \emph{fine-tuning} paradigm, given a train dataset $\mathcal{D}$, the full set of the pretrained weights $\phi_0$ are updated by gradient descent methods to optimize the objective:
\begin{equation}
\max_\phi\sum_{(\mathbf{x}, \mathbf{y})\in \mathcal{D}}\sum_{i=1}^{|\mathbf{y}|}\log {(p_\phi(y_i|\mathbf{x}, \mathbf{y}_{< i}))}.
\end{equation}


\subsection{Prompt-Tuning}

\citet{NEURIPS2020_1457c0d6} showed that \emph{prompt design} is surprisingly effective at adapting multiple downstream tasks to a frozen GPT-3 model, by prepending a natural language task instruction and a few examples to the task input.
\citet{lester2021power} proposed prompt-tuning, in which each prompt is not a text sequence manually designed, but a sequence of continous embeddings that can be optimized by learning from the train dataset $\mathcal{D}$.

Given a PLM $p_\phi$ and an input sequence $\mathbf{x}=\{x_1, ..., x_n\}$, the first thing $p_\phi$ does is to embed $\mathbf{x}$ as a matrix $\mathit{X}\in\mathcal{R}^{n\times e}$, where $e$ is the dimension of the embedding space.
In prompt-tuning, a soft prompt can also be represented as a matrix $\mathit{C}\in\mathcal{R}^{k\times e}$, where $k$ is a hyperparameter that controls the prompt length.
This soft prompt is then concatenated to $\mathit{X}$, forming a single matrix $[\mathit{C};\mathit{X}]\in\mathcal{R}^{(k+n)\times e}$.
This matrix is feed into the frozen PLM (from the bottom layer of the Transformer architecture) to generate the output sequence $\mathbf{y}$:
\begin{equation}
P(\mathbf{y}|\mathbf{x})=\prod_{i=1}^{m}{p_\phi(y_i|[\mathit{C};\mathit{X}], \mathbf{y}_{< i})}.
\end{equation}

During training, PLM parameters are frozen (i.e., $\phi\equiv\phi_0$), while the parameters of $\mathit{C}$ (denoted as $\theta_\mathit{C}$) are updated by:
\begin{equation}
\max_{\theta_\mathit{C}}\sum_{(\mathbf{x}, \mathbf{y})\in \mathcal{D}}\sum_{i=1}^{|\mathbf{y}|}{\log}(p_{\phi_0}(y_i|[\mathit{C};\mathit{X}], \mathbf{y}_{< i})).
\label{eq:prompt_tuning}
\end{equation}

Existing work uses prompt-tuning to address tasks like SuperGLUE~\cite{lester2021power}, while lacking further investigation on its effectiveness for more tasks such as machine translations.


\section{Preliminary Exploration}

\begin{figure}[t]
	\centering
	\includegraphics[width=0.45\textwidth]{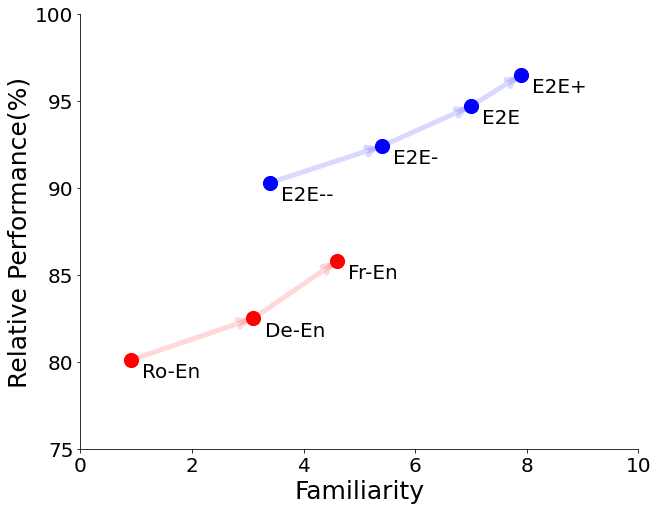}
	\caption{Relative performance and input familiarity of \textcolor[rgb]{0.9,0,0}{machine translation} and \textcolor[rgb]{0,0,0.9}{table-to-text} tasks. 
	It shows the trend that the relative performance of the prompt-tuning rises as the pretrained model's familiarity with the task inputs increases. 
	This trend indicates that alleviating the unfamiliarity by transforming the inputs can help prompt-tuning perform better.}
	\label{fig:rp_fam}
\end{figure}

\begin{table}[t]
\centering
\resizebox{.9\linewidth}{!}{
\begin{tabular}{ll}
\toprule[2pt]
Variant & Input Example \\ \midrule[1.5pt]
E2E & \begin{tabular}[c]{@{}l@{}}name{[}The Punter{]}, food{[}Indian{]}, \\ priceRange{[}cheap{]}\end{tabular} \\ \midrule[.5pt]
E2E$+$ & \begin{tabular}[c]{@{}l@{}}name is The Punter, food is Indian, \\ priceRange is cheap.\end{tabular} \\ \midrule[.5pt]
E2E$-$ & \begin{tabular}[c]{@{}l@{}}nom{[}The Punter{]}, nourriture{[}Indian{]}, \\ gamme de prix{[}cheap{]}\end{tabular} \\ \midrule[.5pt]
E2E$--$ & \begin{tabular}[c]{@{}l@{}}nom{[}Le Punter{]}, nourriture{[}Indienne{]}, \\ gamme de prix{[}pas cher{]}\end{tabular} \\ \midrule[2pt]
\end{tabular}
}
\caption{Variants of E2E inputs.
The $\mathbf{+}$ suffix means that we increase the familiarity by manually transforming the inputs, while $\mathbf{-}$ and $\mathbf{--}$ contra.
}
\label{tab:input_eg}
\end{table}

In this section, we start from an observation that on machine translation tasks there are performance gaps between prompt-tuning and fine-tuning, which are related to unfamiliar inputs (Section \ref{section:unfamiliar}).
We further find that this gap can be controlled by manually transforming the inputs (Section \ref{section:control}), motivating the idea that we can use a trainable module to learn what is the best way to transform the unfamiliar inputs (Section \ref{section:motivation}).


\subsection{Unfamiliar Inputs in Prompt-Tuning}
\label{section:unfamiliar}


We start by investigating the effectiveness of prompt-tuning on three machine translation tasks: Fr-En, De-En and Ro-En.
We use the datasets from WMT 2014 \cite{bojar2014findings} for Fr-En and De-En, and WMT 2016 \cite{bojar2016findings} for Ro-En.
Concretely, we use T5-Large~\citep{2020t5} as the frozen PLM, and measure the gap between \textbf{P}rompt-\textbf{T}uning (PT) and \textbf{F}ine-\textbf{T}uning (FT) based on the \textbf{R}elative \textbf{P}erformance (RP), which is calculated by:
\begin{equation}
\begin{split}
    \mathrm{RP} = \frac{\mathrm{BLEU_{PT}}}{\mathrm{BLEU_{FT}}}.
\end{split}
\end{equation}
Experimental results show that, on all machine translation tasks, prompt-tuning can only achieve 85.8\%\,/\,82.5\%\,/\,80.1\% of the fine-tuning performance, respectively.
This unexpected phenomena leads to a natural follow-up research question: \emph{where does the gap come from?}


Since T5-Large is mostly pretrained on English natural language corpus, a reasonable speculation to the question is that the frozen PLM lacks sufficient knowledge of how to encode the non-English inputs in these three tasks, leading to the large gap between prompt-tuning and fine-tuning.
In other words, \emph{the frozen PLM is unfamiliar with the inputs}.
To make it more clear, we quantify the \textbf{Familiarity} (denoted by Fam) of a task's inputs to a frozen PLM using the bi-gram frequencies in the pretraining corpus.
Concretely, we define the Familiarity of task inputs as:
\begin{equation}
\begin{split}
    \mathrm{Fam} = \frac{1}{|\mathbf{X}|}\sum_{\mathbf{x}\in{\mathbf{X}}}  \sum_{i=1}^{n-1} \frac{\log{\mathrm{BiC}(\mathbf{x}_{i:i+1})}}{n-1},
\end{split}
\end{equation}
where $\mathbf{x}$ is a token sequence from all inputs $\mathbf{X}$ in the task, and $\mathrm{BiC}(\mathbf{x}_{i:i+1})$ returns the bi-gram counts of $\mathbf{x}_{i:i+1}$.
The \textcolor[rgb]{0.9,0,0}{red} points in Figure~\ref{fig:rp_fam} shows the RP of translation tasks with different input familiarities.
It shows the trend that the more familiar with inputs, the higher RP the prompt-tuning achieves, which implies that the performance gap partly comes from the unfamiliarity.
Based on this observation, we suppose that \textbf{alleviating the unfamiliarity by transforming the inputs can help prompt-tuning perform better.}

\subsection{Manually Transforming Inputs}
\label{section:control}

Next step, to evaluate the effectiveness of transforming inputs, we conduct some controllable experiments based on a table-to-text task E2E~\cite{novikova2017e2e}.
The inputs of E2E are synthetic linearized tables, which are easy to manually transform.
We consider two kinds of input transformations: more familiar transformation (denote with suffix $\mathbf{+}$) that changes the inputs more like natural language, and less familiar transformation  (denote with suffix $\mathbf{-}$ and $\mathbf{--}$) that makes the bi-gram counts of input lower.
Details of these transformations are listed in the Appendix~\ref{sec:appendix_a}.
Some input examples are shown in Table~\ref{tab:input_eg}.

As shown in Figure~\ref{fig:rp_fam} (painted in \textcolor[rgb]{0,0,0.9}{blue}), the task with more familiar inputs (E2E$\mathbf{+}$) achieve better performance than the original task while the tasks with less familiar inputs (E2E$\mathbf{-}$ and E2E$\mathbf{--}$) perform contra.
We also conduct another controllable experiment on logic-to-text tasks and observe the similar performances (detailed in Appendix~\ref{sec:appendix_a}).
This trend is similar to it in machine translation tasks, which assists the speculation before.
Furthermore, observing the gain from E2E to E2E$\mathbf{+}$, it indicates that transforming inputs is a promising way to alleviate the unfamiliarity and improve the prompt-tuning.

\subsection{Motivation}
\label{section:motivation}
As discussed before that manually transforming the inputs can benefit the prompt-tuning, but apparently, it is inefficient and error-prone for non-synthetic tasks such as translations.
Moreover, separating the transforming processes with prompt-tuning could cause incompatibility between two modules.

Based on these concerns, our motivation is to design a transforming method that meets two requirements: first, it should be universally applicable for different tasks and pretrained backbones.
Second, it could be optimized automatically and jointly with prompt-tuning.
Therefore, we proposed to \textbf{utilize a neural-based adapter to transform the continuous input embeddings}.

\section{Input-Tuning}\label{sec:method}

Based on our motivation, we extend the prompt-tuning with a neural-based adapter to transform the input embeddings, namely \emph{input-tuning}.
Figure~\ref{fig:input_tuning} illustrates the basic idea of input-tuning.
Given a frozen PLM $p_{\phi_0}$ and an input sequence $\mathbf{x}$, input-tuning generates the output sequence $\mathbf{y}$ by:
\begin{equation}
P(\mathbf{y}|\mathbf{x})=\prod_{i=1}^{m}{p_{\phi_0}(y_i|\mathcal{F}(\mathbf{x}), \mathbf{y}_{< i})}
\end{equation}
where $\mathcal{F}(\cdot)$ is a learnable function deciding how $\mathbf{x}$ should be ``tuned'' to activate $p_{\phi_0}$ to generate proper $\mathbf{y}$.
Prompt-tuning (Equation \ref{eq:prompt_tuning}) can be regarded as a special case of input-tuning, where $\mathcal{F}(\cdot)$ is defined as the concatenation of $\mathit{C}$ (the soft prompt) and $\mathit{X}$ (the embedding matrix of $\mathbf{x}$).

\begin{figure}[t]
	\centering
	\includegraphics[width=0.22\textwidth]{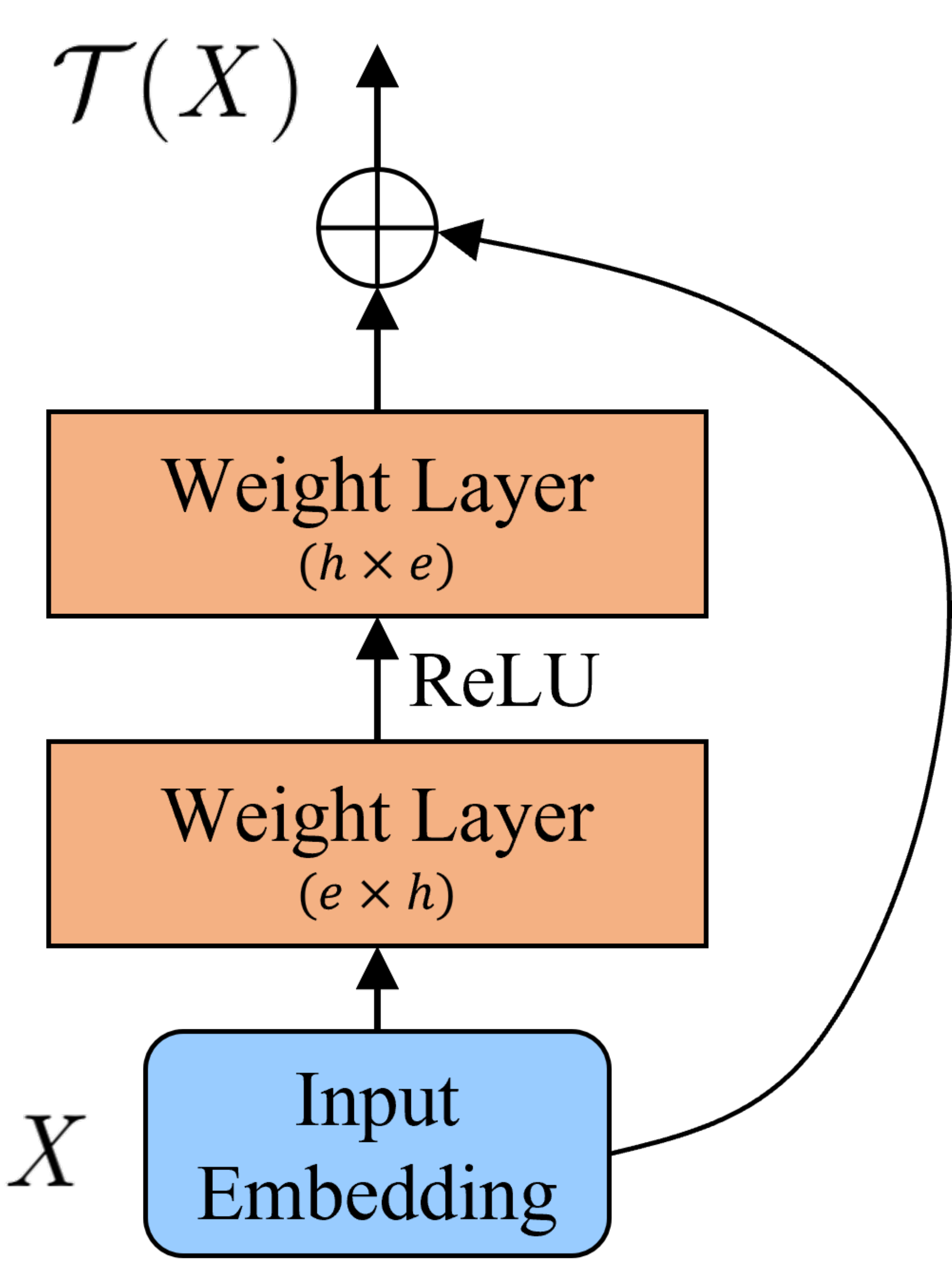}
	\caption{The illustration of our Input-Adapter.}
	\label{fig:input_adapter}
\end{figure}

In this paper, we model $\mathcal{F}(\cdot)$ as:
\begin{equation}
\mathcal{F}(\mathbf{x}) = [\mathit{C}; \mathcal{T}(\mathit{X})],
\end{equation}
\vspace{-0.7cm}
\begin{equation}
\begin{split}
\mathcal{T}(\mathit{X})_{i,*}=
&\mathit{X}+\sigma(\mathit{X}_{i,*}\mathbf{W_1})\mathbf{W_2}, \\
&\text{for each $i=1,...,n.$}
\end{split}
\end{equation}
Here $\sigma$ is an element-wise nonlinear activation function (ReLU).
$\mathbf{W_1}$ and $\mathbf{W_2}$ are learnable matrices.
We denote $\mathcal{T}(\cdot)$ as an \emph{input-adapter}, since it plays the role that adapts the surface representations of $\mathbf{x}$ to better utilize the frozen PLM.
Figure~\ref{fig:input_adapter} illustrates this module.

\paragraph{Initialization.}
We initialize the soft prompt $\mathit{C}$ by randomly picking from pretrained embeddings, as \citet{lester2021power} claimed that it is much better than totally random initialization.
For parameters of the input-adapter, we initialize $\mathbf{W_2}$ as a zero matrix so that the $\mathcal{T}(\mathit{X})$ will be the same with pretrained embeddings at the beginning of training.


\paragraph{Token-wise or Sequence-wise.}
Input-adapter tunes the surface representation of each token in $\mathbf{x}$ respectively, without taking sequence contexts into consideration (e.g., model $\mathcal{T}(\mathit{X})$ using RNNs or Transformers).
The key reason of this design choice is to prevent the input-adapter dominating the frozen PLM.
Concretely, as RNNs and Transformers are already good sequence-to-sequence learners, using them to implement the input-adapter may lead to the result that the sequence-to-sequence task is mainly addressed in the input-adapter while the frozen PLM contributes little.
Therefore, we chose a simple token-wise MLP to implement the input-adapter, which is far from sufficient for sequence-to-sequence learning, thus activating the frozen PLM better.
Our experimental results also support this design choice (Section \ref{sec:diff_input_adapters}).

\section{Experiments}
\label{section:experiments}

\begin{table}[t]
\centering
\resizebox{1.\linewidth}{!}{
\begin{tabular}{ c | c c c c}
\toprule[1.5pt]
 & \multirow{2}*{\textbf{Data Size}} & \multirow{2}*{\textbf{Vocab Size}} & \textbf{Input Len} & \textbf{Output Len} \\
 & & & \textbf{(Avg)} & \textbf{(Avg)} \\
\midrule[1pt]
E2E & 47.4K & 3.0K & 30.3 & 21.7 \\
DART & 81.6K & 33.2K & 30.2 & 21.1 \\
ToTTo & 136.1K & 136K & 151.3 & 18.0 \\
ONR & 13.7K & 1.3K & 56.9 & 9.3 \\
Fr-En & $\sim$ 2M & 373K & 28.8 & 27.5 \\
De-En & $\sim$ 2M & 586K & 26.2 & 27.4 \\
Ro-En & $\sim$ 400K & 169K & 26.8 & 26.8 \\
\bottomrule[1.5pt]
\end{tabular}
}
\caption{
Task statistics.
}
\label{tab:t2t_dataset_overview}
\end{table}

\begin{table*}
\renewcommand\arraystretch{1.2}
\Huge
\centering
\resizebox{0.99\linewidth}{!}{
\begin{tabular}{lcccccccccccccccc}
\toprule[3pt]
 & \multicolumn{5}{c|}{E2E} & \multicolumn{6}{c|}{DART} & \multicolumn{5}{c}{ONR} \\
 & BLEU & NIST & MET & R-L & \multicolumn{1}{c|}{CIDEr} & BLEU & MET & TER↓ & Mov & BERT & \multicolumn{1}{c|}{BLRT} & BLEU & NIST & MET & R-L & CIDEr \\
\midrule[1.5pt]
\multicolumn{6}{c}{} & \multicolumn{6}{c}{T5-Large} \\
Prompt-Tuning & 64.5 & 8.35 & 44.1 & 67.3 & \multicolumn{1}{c|}{2.23} & 46.9 & \underline{0.39} & 0.47 & 0.51 & \underline{\textbf{0.95}} & \multicolumn{1}{c|}{0.45} & 65.3 & 10.10 & 46.9 & 78.4 & 2.76 \\
\graycell Input-Tuning & \underline{\textbf{68.7}} & \underline{\textbf{8.74}} & \underline{46.1} & \underline{70.7} & \multicolumn{1}{c|}{2.42} & \underline{48.3} & \underline{0.39} & \underline{0.46} & \underline{0.52} & \underline{\textbf{0.95}} & \multicolumn{1}{c|}{\underline{0.47}} & \underline{\textbf{70.2}} & \underline{\textbf{10.50}} & \underline{49.1} & \underline{80.6} & \underline{2.84} \\
\cmidrule(l){1-17}
Fine-Tuning & 68.1 & 8.67 & \textbf{46.6} & \textbf{71.3} & \multicolumn{1}{c|}{2.44} & \textbf{49.2} & \textbf{0.40} & \textbf{0.44} & \textbf{0.53} & \textbf{0.95} & \multicolumn{1}{c|}{\textbf{0.48}} & 69.4 & 10.41 & \textbf{49.4} & \textbf{81.4} & \textbf{2.87} \\
\midrule[1.5pt]

\multicolumn{6}{c}{} & \multicolumn{6}{c}{GPT2-Large} \\
Prompt-Tuning & 65.8 & 8.51 & 44.4 & 68.2 & \multicolumn{1}{c|}{2.23} & 43.5 & 0.37 & 0.51 & 0.48 & \underline{\textbf{0.94}} & \multicolumn{1}{c|}{0.39} & 64.2 & 9.71 & 44.3 & 75.6 & 2.62 \\
\graycell Input-Tuning &  \underline{68.2} & \underline{\textbf{8.79}} &  \underline{45.3} & \underline{\textbf{70.2}} & \multicolumn{1}{c|}{\underline{2.34}} & \underline{46.4} &  \underline{0.38} & \underline{0.50} & \underline{0.49} & \underline{\textbf{0.94}} & \multicolumn{1}{c|}{\underline{\textbf{0.42}}} & \underline{\textbf{68.4}} & \underline{\textbf{10.01}} & \underline{\textbf{46.9}} & \underline{\textbf{78.5}} & \underline{\textbf{2.75}} \\
\cmidrule(l){1-17}
Fine-Tuning & \textbf{68.5} & 8.78 & \textbf{46.0} & 69.9 & \multicolumn{1}{c|}{\textbf{2.45}} & \textbf{47.0} & \textbf{0.39} & \textbf{0.46} & \textbf{0.51} & \textbf{0.94} & \multicolumn{1}{c|}{0.40} & 67.0 & 9.58 & 45.7 & 77.6 & 2.67 \\
\bottomrule[3pt]
\end{tabular}
}
\caption{
Results of table-to-text tasks \textbf{E2E} (left), \textbf{DART} (middle) and logic-to-text task \textbf{ONR} (right).
The \textbf{bold scores} are the best results among all methods, and the \underline{underlined scores} represent the best results in prompt-tuning and input-tuning.
The fine-tuning results of GPT2-Large on E2E and DART are reported by \citet{li2021prefix}.
On all three tasks, our input-tuning significantly outperforms the prompt-tuning. 
Impressively, it can achieve comparable or even better performance than fine-tuning on these tasks. 
}
\label{tab:basic}
\end{table*}

\subsection{Experimental Setup}\label{sec:exp_setup}

\paragraph{Tasks and Metrics} \label{sec:tasks}
We evaluate on seven NLG tasks: three table-to-text tasks, one logic-to-text task, and three machine translation tasks.
Table~\ref{tab:t2t_dataset_overview} shows the statistics of these datasets.

The \emph{table-to-text} tasks are: E2E~\cite{novikova2017e2e},  DART~\cite{nan2020dart}, and ToTTo~\cite{parikh2020ToTTo}.
E2E is relatively simpler than the other two tasks, as it only has 1 domain (i.e., restaurant reviews).
DART and ToTTo are both open-domain, using open-domain tables from Wikipedia.
ToTTo is the most challenging task among them, as it increases the generalization challenge by reducing the vocabulary overlap between train set and dev/test set.

The \emph{logic-to-text} task is ONR (OverNight-Reverse), which is constructed based on the text-to-logic benchmark Overnight~\cite{wang2015building} (detailed in Appendix~\ref{sec:appendix_b}).

The \emph{Machine translation} tasks are: French to English (Fr-En), German to English (De-En) and Romanian to English (Ro-En).
We use the parallel corpus from version 7 of Europarl corpus \cite{koehn2005europarl} as the training data for all three translation tasks.
For Fr-En and De-En, we take WMT $\mathrm{newstest2013}$ $\mathrm{newstest2014}$ \cite{bojar2014findings} as the development sets and test sets, respectively.
For Ro-En, we use the development and test sets from WMT 2016 \cite{bojar2016findings}.

For every tasks, we report the official evaluation metrics, including BLEU \cite{papineni2002bleu}, NIST \cite{belz2006nist}, METEOR \cite{lavie2007meteor}, ROUGE-L \cite{lin2004rouge}, CIDEr \cite{vedantam2015cider}, TER \cite{snover2005ter}, MoverScore \cite{zhao2019moverscore}, BERTScore \cite{zhang2019bertscore}, BLEURT \cite{sellam2020bleurt} and PARENT \cite{dhingra2019parent}.

\paragraph{Backbones and Hyperparameters}

We choose T5-Large ($\sim$770M parameters) and GPT2-Large (774M parameters) as our backbones.
Our implementation is based on the HuggingFace Transformer models \cite{wolf2020transformers}.
The training hyperparameters mainly include the learning rate (1e-5$\sim$5e-3), weight decay (1e-1 or 1e-2) and learning rate scheduler (linear or constant).
The evaluation hyperparameters mainly include the beam size (1 or 5).
All experimental results are produced under the same hyperparameter searching strategy except the cited results.
The default prompt length is 100 (the best prompt length indicated by \citet{lester2021power}) and the hidden layer dimensions in the input-adapter are two times of the embedding dimensions. 
The reparameterization trick used by \citet{li2021prefix} is applied.
More hyperparameter details are listed in the Appendix~\ref{sec:appendix_c}.


\begin{table}[]
\renewcommand\arraystretch{1.1}
\large
\centering
\resizebox{1.\linewidth}{!}{
\begin{tabular}{lccc}
\toprule[1.5pt]
 & \multicolumn{3}{c}{ToTTo (PARENT on Dev Set)} \\
 & \multicolumn{1}{c}{Overall} & \multicolumn{1}{c}{Overlap} & \multicolumn{1}{c}{Non-Overlap} \\
\midrule[1pt]
Prompt-Tuning & 56.0 & 59.3 & 52.9 \\
\graycell Input-Tuning & \underline{58.0} & \underline{61.3} & \underline{54.7} \\
\midrule[0.5pt]
Fine-Tuning & \textbf{59.8} & \textbf{63.4} & \textbf{56.3} \\
\bottomrule[1.5pt]
\end{tabular}
}
\caption{PARENT scores on \textbf{ToTTo}, with T5-Large as the backbone.
Input-tuning outperforms prompt-tuning, and the relative gains on Overlap and Non-Overlap subsets are on par (2.0 and 1.8, respectively), showing that it maintains a stable generalization ability.}
\label{tab:ToTTo}
\end{table}

\subsection{Main Results}

Table~\ref{tab:basic}, \ref{tab:ToTTo} and \ref{tab:translation} show the main results on seven NLG tasks.
In general, our input-tuning method notably outperforms prompt-tuning on all these tasks. 
Moreover, on E2E, DART and ONR tasks (shown in Table~\ref{tab:basic}), input-tuning achieves comparable or even better performance than fine-tuning.

\begin{table}[t]
\renewcommand\arraystretch{1.1}
\large
\centering
\resizebox{1.\linewidth}{!}{
\begin{tabular}{lcccccc}
\toprule[1.5pt]
 & \multicolumn{6}{c}{Machine Translation (BLEU)} \\
 & \multicolumn{2}{c}{Fr-En} & \multicolumn{2}{c}{De-En} & \multicolumn{2}{c}{Ro-En} \\
 \cmidrule(l){2-7} 
 & Dev & Test & Dev & Test & Dev & Test \\
\midrule[1pt]
Prompt-Tuning & 26.9 & 29.6 & 24.4 & 24.5 & 26.7 & 26.5 \\
\graycell Input-Tuning & \underline{28.5} & \underline{31.7} & \underline{25.9} & \underline{26.2} & \underline{31.3} & \underline{30.4} \\
\midrule[0.5pt]
Fine-Tuning & \textbf{30.9} & \textbf{34.5} & \textbf{29.0} & \textbf{29.7} & \textbf{34.4} & \textbf{33.1} \\
\bottomrule[1.5pt]
\end{tabular}
}
\caption{BLEU scores on \textbf{Machine Translation} tasks(Fr/De/Ro-En), with T5-Large as the backbone. 
}
\label{tab:translation}
\end{table}

\paragraph{Table-to-Text}
With an additional lightweight input-adapter, the input-tuning notably outperforms prompt-tuning on three table-to-text tasks including E2E, DART and ToTTo (shown in Table~\ref{tab:basic} and \ref{tab:ToTTo}). 
Surprisingly, it can be comparable or even better than fine-tuning on E2E and DART tasks. 
For the ToTTo task, input-tuning still outperforms prompt-tuning and slightly lags fine-tuning, and the relative gains on Overlap and Non-Overlap subsets are on par (2.0 and 1.8, respectively), showing that it maintains a stable generalization ability.
These results show that a simple input-adapter can bring significant benefits.

\paragraph{Logic-to-Text} 
Unlike table-to-text tasks, the inputs of logic-to-text tasks contain more nested and complex logic forms.
On ONR task, the input-tuning also apparently outperforms prompt-tuning, and even surprisingly outperforms fine-tuning (shown in Table~\ref{tab:basic}). 
These results indicate that the input-adapter can handle different input formats and keep a stable performance.

\paragraph{Machine Translation}
Compared with table-to-text and logic-to-text, the translation tasks are much more challenging as they contain larger vocabularies and more training data, making them much harder for lightweight models.
Table~\ref{tab:translation} shows that on all three translation tasks, input-tuning significantly closes the gap between prompt-tuning and fine-tuning. 
It demonstrates that input-tuning consistently outperforms prompt-tuning even on more challenging tasks.

\section{Analysis}

In this section, we further explore the effectiveness of input-tuning under different settings.
Section \ref{sec:diff_backbones} analyzes the performance with different PLM backbones.
Section \ref{sec:low_resource} explores the different data scale settings.
Section \ref{sec:diff_input_adapters} discusses the different designs of input-tuning.
And Section \ref{sec:diff_prompt_length} shows the performance with different lengths of soft prompts.


\subsection{Different Backbones} \label{sec:diff_backbones}
We evaluate the effectiveness on different backbones from two aspects: 
backbone architectures (GPT2-Large and T5-Large) and backbone sizes (T5-Small/Base/Large).
Note that as the input-adapter only modifies the input embeddings, it can be universally applied to different types of backbones with word embeddings as input.

\paragraph{Backbone Architectures}
The results in Table \ref{tab:basic} show that although the fine-tuning performance of GPT2-Large and T5-Large vary, 
input-tuning maintains better results than prompt-tuning. 
It implies the potential of input-tuning to be effective for a variety of model architectures.

\begin{figure}[t] 
    \centering 
    \includegraphics[width=0.47\textwidth]{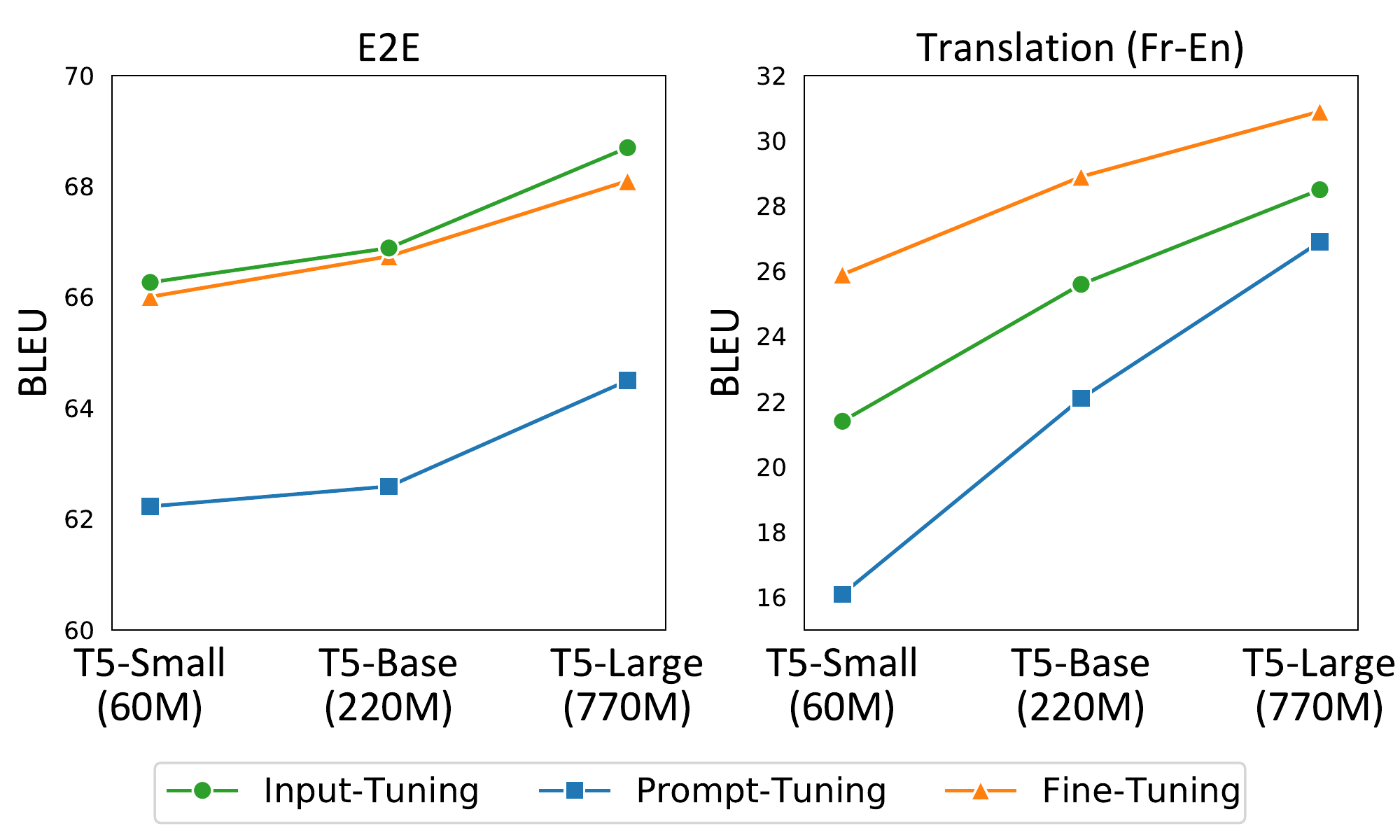}
    \caption{
    The performance with different backbone sizes (T5-Small/Base/Large) on E2E task and translation task (Fr-En dev set). 
    The input-tuning stably outperforms prompt-tuning and benefits from the increasing of backbone size.
    }
    \label{fig:model_size} 
\end{figure}

\paragraph{Backbone Sizes}
As shown in Figure~\ref{fig:model_size}, the input-tuning stably outperforms prompt-tuning with different sizes of T5 (Small, Base and Large).
Furthermore, compared with fine-tuning, the input-tuning benefits more from the increasing of backbone size, indicating that it can be more effective with larger pretrained models.

\begin{figure}[t] 
    \centering 
    \includegraphics[width=0.47\textwidth]{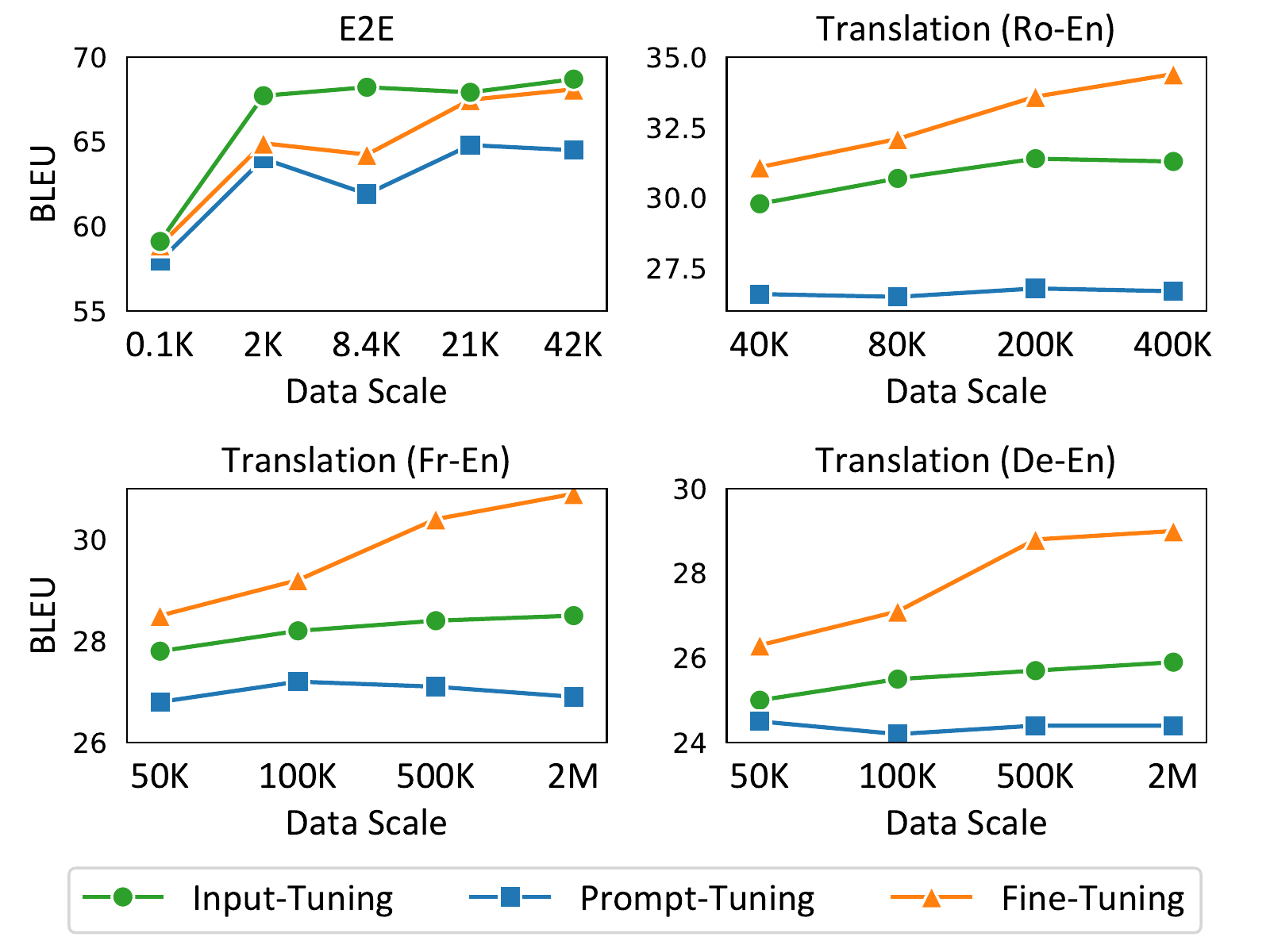}
    \caption{
    Performance under different data scales on E2E and three translation tasks (dev sets).
    The input-tuning stably performs better than prompt-tuning and prefers low-resource scenarios.
    } 
    \label{fig:low_resource} 
\end{figure}

\subsection{Different Data Scales} \label{sec:low_resource}
To explore the impact of data scales, we conduct experiments on different scales of E2E and translation tasks.
The experimental results in Figure~\ref{fig:low_resource} show that the input-tuning stably performs better than prompt-tuning under different data scales. 
Compared with fine-tuning, the input-tuning consistently outperforms it in E2E, which is a relative low-resource task compared with translations.
And for translation tasks, the gaps between input-tuning and fine-tuning are closed with the data scales decreasing.
These results indicate that the input-tuning prefers low-resource scenarios.

\begin{figure}[t]
    \centering
    \includegraphics[width=0.47\textwidth]{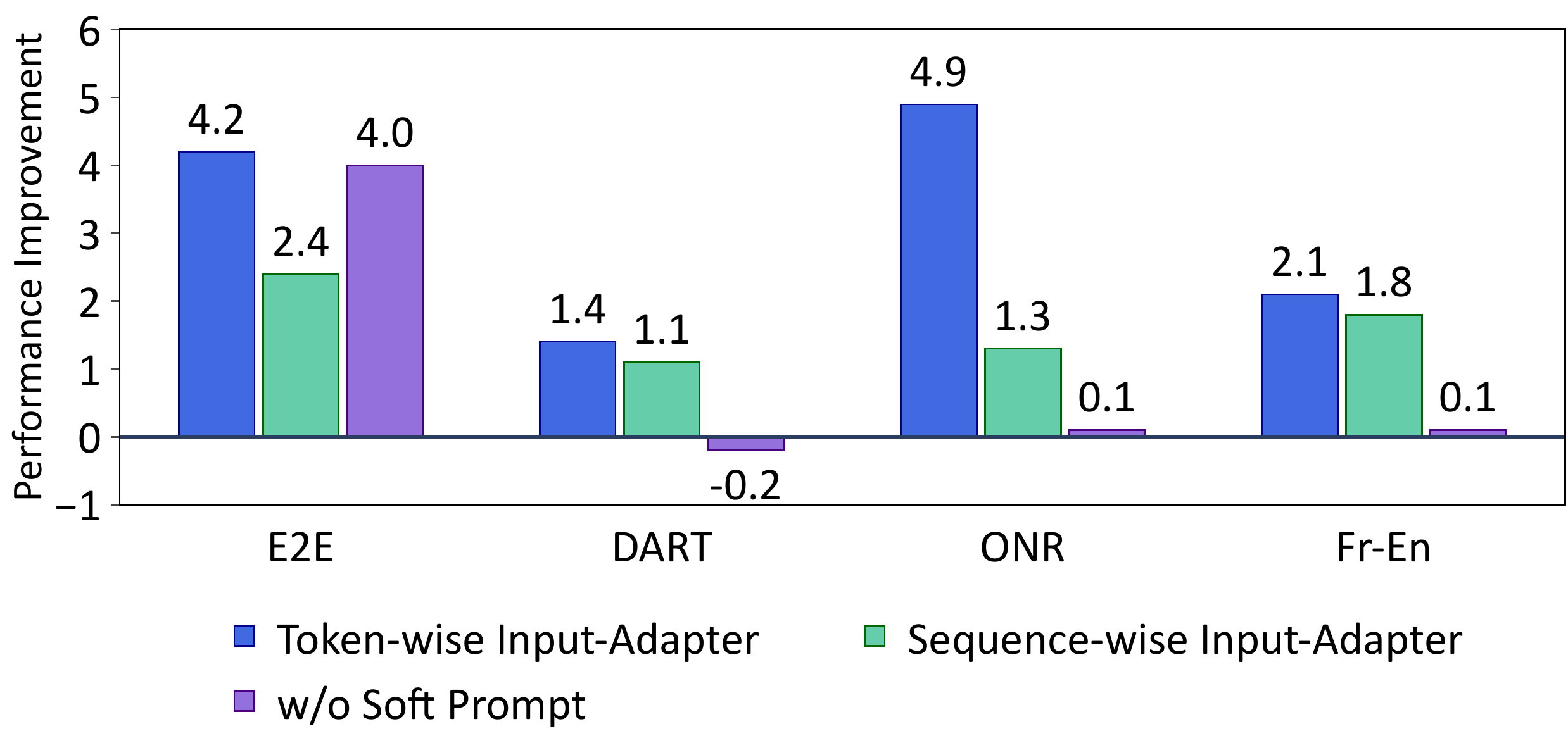}
    \caption{
    The improvements of input-tuning (w.r.t. prompt-tuning) with different designs.
    We take BLEU scores for these four tasks.
    It shows that a simple token-wise input-adapter can activate the frozen PLM better, and the soft prompts are indispensable in input-tuning.
    }
    \label{fig:ablation_4_tasks}
\end{figure}




\subsection{Different Designs of Input-Tuning} \label{sec:diff_input_adapters}
Besides the original design described in Section~\ref{sec:method} (the token-level input-adapter with soft prompts), we consider another two designs: the sequence-wise input-adapter with soft prompts, and the token-level input-adapter without soft prompts.
Figure~\ref{fig:ablation_4_tasks} shows the results on four tasks\footnote{The results for more tasks are shown in the Appendix~\ref{sec:appendix_d}.}.

\paragraph{Sequence-wise Input-Adapter}
The sequence-wise input-adapter is implemented with a self-attention layer.
As shown in Figure~\ref{fig:ablation_4_tasks} (\textcolor[rgb]{0,0.5,0}{green bars}), it always performs worse than the original one with a token-wise input-adapter (\textcolor[rgb]{0,0,0.7}{blue bars}), supporting our speculation in Section~\ref{sec:method} that a stronger adapter may dominate the pretrained model and lead to the performance loss while a simple and lightweight module can activate the frozen PLM better.

\paragraph{Without Soft Prompts}
The \textcolor[rgb]{0.4,0.1,0.9}{purple bars} in Figure~\ref{fig:ablation_4_tasks} shows the results of input-tuning w/o soft prompts.
With only a token-level input-adapter, the input-tuning loses much performance and sometimes even lags behind the prompt-tuning, indicating that soft prompts are indispensable in the input-tuning.

\subsection{Different Lengths of Soft Prompts} \label{sec:diff_prompt_length}
As soft prompts are important for input-tuning, here we further explore the impact of different prompt lengths.
We select prompt lengths from 1 to 200 and conduct experiments on E2E and Fr-En tasks.

\begin{figure}[t] 
    \centering 
    \includegraphics[width=0.48\textwidth]{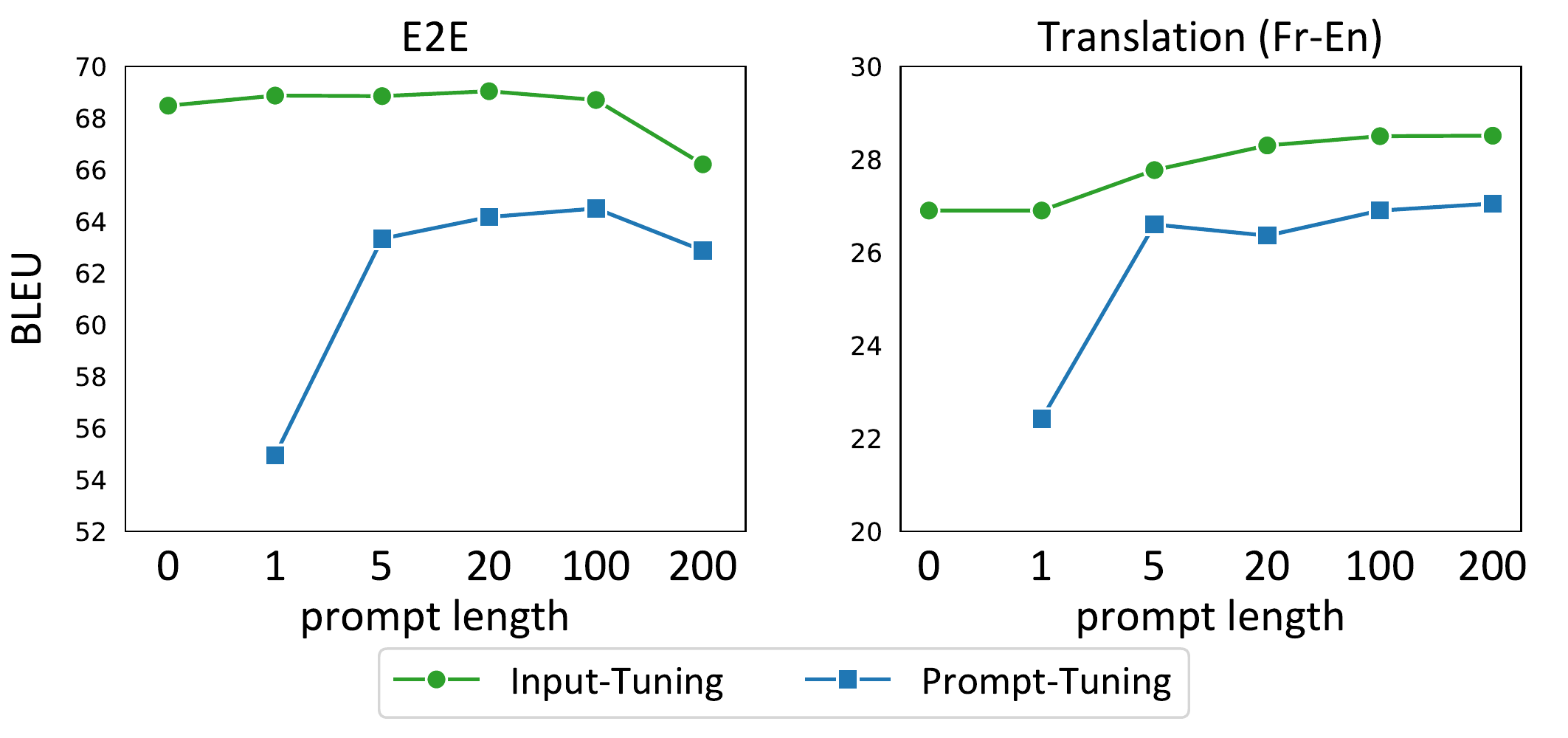}
    \caption{
    The performance of input-tuning and prompt-tuning with different prompt lengths.
    The 0 prompt length means that there is only an input-adapter (see Section~\ref{sec:diff_input_adapters}).
    The input-tuning performs better than prompt-tuning with all prompt lengths.
    } 
    \label{fig:prompt_length} 
\end{figure}

As shown in Figure~\ref{fig:prompt_length}, the input-tuning is better than prompt-tuning with all prompt lengths. 
Moreover, it seems that different tasks prefer different input lengths.
For the relative simple task E2E, it shows that a short soft prompt is enough while the longer prompt length may hurt the performance.
For the more challenging task Fr-En, both input-tuning and prompt-tuning prefers longer inputs.

\section {Related Work}

\paragraph{Parameter-Efficient Fine-Tuning}
Despite the strong ability of large-scale pretrained language models \cite{radford2018improving, devlin2019bert, liu2019roberta, raffel2020exploring}, tuning hundreds of billions of parameters is extremely time- and space-consuming. Therefore, parameter-efficient fine-tuning recently attracted more attention. 
There are two series of related works: invasive methods and non-invasive methods.
Invasive methods, which are built on a strong assumption that the inner structure (e.g., self-attention and feed-forward layers) of the PLM can be modified, includes Prefix-Tuning~\cite{li2021prefix}, Bitfit~\cite{ben2021bitfit}, Child-Tuning~\cite{xu2021raise}, P-Tuning v2~\cite{liu2021p-v2}, LoRA~\cite{hu2021lora}, UnifiedSKG~\cite{xie2022unifiedskg} and Adapter-based models~\cite{rebuffi2017learning, houlsby2019parameter, lin2020exploring, he2021effectiveness, pfeiffer2021adapterfusion}.
Non-invasive methods, which only modify input embeddings and regard the inner structure as a black box, mostly are prompting methods (including our Input-Tuning).

\paragraph{Prompting}
Prompting means prepending instructions or a few examples to the task input and generating the output from the PLM.
Recently, various kinds of hard prompt designing and searching methods were proposed~\cite{jiang2020can, shin2020autoprompt, schick2020s, gao2020making}. 
To overcome the shortcomings of hand-crafting, some works optimize continuous soft prompts such as prompt-tuning~\cite{lester2021power}, P-Tuning~\cite{liu2021gpt}, Ppt~\cite{gu2021ppt} and SPoT~\cite{vu2021spot}. 
Recently, \citet{sun2022black} and \citet{diao2022black} explored black-box tuning, which trains the prompt without gradient back-propagation.
Compared with these work, we are the first to study the effectiveness of soft prompts from the view of task inputs.

\section{Conclusion}
In this paper, we revealed that input unfamiliarity is an essential ingredient affecting the performance of prompt-tuning, and proposed input-tuning to alleviate this unfamiliarity with a neural-based input-adapter.
We conducted experiments on seven NLG tasks with thorough analysis, showing that input-tuning performs stably better than prompt-tuning.
\bibliography{anthology,custom}
\bibliographystyle{acl_natbib}

\appendix

\vspace{24 pt}
This is the Appendix for the paper: 
``Input-Tuning: Adapting Unfamiliar Inputs to Frozen Pretrained Models''.

\section{Preliminary Exploration}\label{sec:appendix_a}

Here we show more details in our preliminary exploration.

\subsection{Manually Transforming Inputs}

We take a table-to-text task E2E and logic-to-text task ONR.
Details of these tasks are shown in Section~\ref{sec:exp_setup}.
Here we describe how we transform the inputs of E2E and ONR manually.

\paragraph{E2E$+$} Replace the left brackets $[$ with $is$, remove the right brackets, and add a period at the end of the sentence.

\paragraph{E2E$-$} Translate the attributes into French.

\paragraph{E2E$--$} Translate both the attributes and the values into French.

\paragraph{ONR$+$} We use the canonical utterances for every logic forms.

\paragraph{ONR$-$} Convert the function names into Chinese Pinyin.

The examples of E2E inputs variants are listed in Table~\ref{tab:input_eg}, and here we show the variants of ONR inputs in Table~\ref{tab:input_eg_onr}.

\begin{table}[htbp]
\centering
\resizebox{.9\linewidth}{!}{
\begin{tabular}{ll}
\toprule[2pt]
Variant & Input Example \\ \midrule[1.5pt]
ONR & \begin{tabular}[c]{@{}l@{}}call listvalue (call getproperty \\ en.person.alice (string birthplace))\end{tabular} \\ \midrule[.5pt]
ONR$+$ & \begin{tabular}[c]{@{}l@{}}city that is birthplace of alice and \\ that is birthplace of alice\end{tabular} \\ \midrule[.5pt]
ONR$-$ & \begin{tabular}[c]{@{}l@{}}diaoyong lieju (diaoyong dedao \\ en.person.alice (zifuchuan birthplace))\end{tabular} \\ \midrule[2pt]
\end{tabular}
}
\caption{Variants of ONR inputs.
}
\label{tab:input_eg_onr}
\end{table}

\subsection{More Controllable Experiments}

We list the relative performance of prompt-tuning and input familiarity in Table~\ref{tab:input_result_onr}.

\begin{table}[htbp]
\centering
\resizebox{.8\linewidth}{!}{
\begin{tabular}{lcc}
\toprule[2pt]
Variant & \begin{tabular}[c]{@{}c@{}}Relative\\ Performance(\%)\end{tabular} & Familiarity \\ \midrule[1.5pt]
ONR & 94.1 & 3.2 \\ \midrule[0.5pt]
ONR$+$ & 97.1 & 11.8 \\ \midrule[0.5pt]
ONR$-$ & 89.3 & 1.1 \\ \midrule[2pt]
\end{tabular}
}
\caption{Relative performance (w.r.t. fine-tuning) and input familiarity of different ONR variants.
}
\label{tab:input_result_onr}
\end{table}

It shows the similar performances with E2E variants that the ONR$+$, which with more familiar inputs, achieves better performance than original task ONR while ONR$-$ performs contra.
These results support our intuition that alleviating the unfamiliarity by transforming the inputs can help prompt-tuning perform better.

\section{Construction of ONR}\label{sec:appendix_b}

In this section, we present how we construct the ONR (\textbf{O}ver\textbf{N}ight-\textbf{R}everse) task based on the text-to-logic benchmark Overnight~\cite{wang2015building}.

The original task of Overnight dataset is semantic parsing, which converts natural language inputs into logic forms in 8 different domains. In our experiment, we mix all the domains and swap the input and output (i.e. use logic forms as inputs and natural language sentences as outputs), and merge different NL outputs with the same logic form input as its candidate outputs, in order to compute metrics (such as BLEU) more precisely. We then shuffle and divide them as new train (9030), dev (2262) and test (2390) set, using the original proportion as overnight. For the train set, we divide multiple candidates as multiple training data. For the test set, we use identical logic forms who have 5 candidate corresponding NL labels.


\section{Hyperparameters}\label{sec:appendix_c}
In this section, we list the detailed hyperparameters for input-tuning used in our main experiments. The total training steps are 100K steps, and we save the checkpoints and evaluate for every 10k steps. The length penalty for evaluation is 1.0. The other details are listed in Table~\ref{tab:hyperparameters}.

\begin{table*}[t]
\centering
\resizebox{.95\linewidth}{!}{
\begin{tabular}{ccccccccc}
\toprule[2pt]
Task & Backbone & batch size & learning rate & weight decay & learning rate scheduler & warmup ratio & beam size & no\_repeat\_ngram\_size \\ \midrule[1.5pt]
\multirow{2}{*}{E2E} & T5-Large & 16 & 5e-4 & 1e-2 & constant & - & 1 & 3 \\ \cmidrule(l){2-9} 
 & GPT2-Large & 8 & 1e-3 & 1e-1 & warmup & 0.1 & 5 & 0 \\ \midrule[0.5pt]
\multirow{2}{*}{DART} & T5-Large & 8 & 5e-5 & 1e-2 & constant & - & 5 & 3 \\ \cmidrule(l){2-9} 
 & GPT2-Large & 8 & 5e-4 & 1e-1 & warmup & 0.1 & 1 & 0 \\ \midrule[0.5pt]
\multirow{2}{*}{ONR} & T5-Large & 8 & 5e-4 & 1e-2 & constant & - & 1 & 3 \\ \cmidrule(l){2-9} 
 & GPT2-Large & 8 & 1e-3 & 1e-1 & warmup & 0.1 & 5 & 0 \\ \midrule[0.5pt]
ToTTo & T5-Large & 8 & 5e-4 & 1e-2 & constant & - & 1 & 3 \\ \midrule[0.5pt]
Fr-En & T5-Large & 16 & 1e-4 & 1e-2 & constant & - & 5 & 0 \\ \midrule[0.5pt]
De-En & T5-Large & 16 & 1e-4 & 1e-2 & constant & - & 5 & 0 \\ \midrule[0.5pt]
Ro-En & T5-Large & 16 & 1e-4 & 1e-2 & constant & - & 5 & 0 \\ \midrule[1.5pt]
\end{tabular}
}
\caption{
Hyperparameters for input-tuning in different tasks.
}
\label{tab:hyperparameters}
\end{table*}

\section{More Results for Different Designs of Input-Tuning}\label{sec:appendix_d}

Here we show the results on all seven tasks for different designs of input-tuning, as a supplementary to Section~\ref{sec:diff_input_adapters}.

\begin{figure*}[t]
    \centering
    \includegraphics[width=1\textwidth]{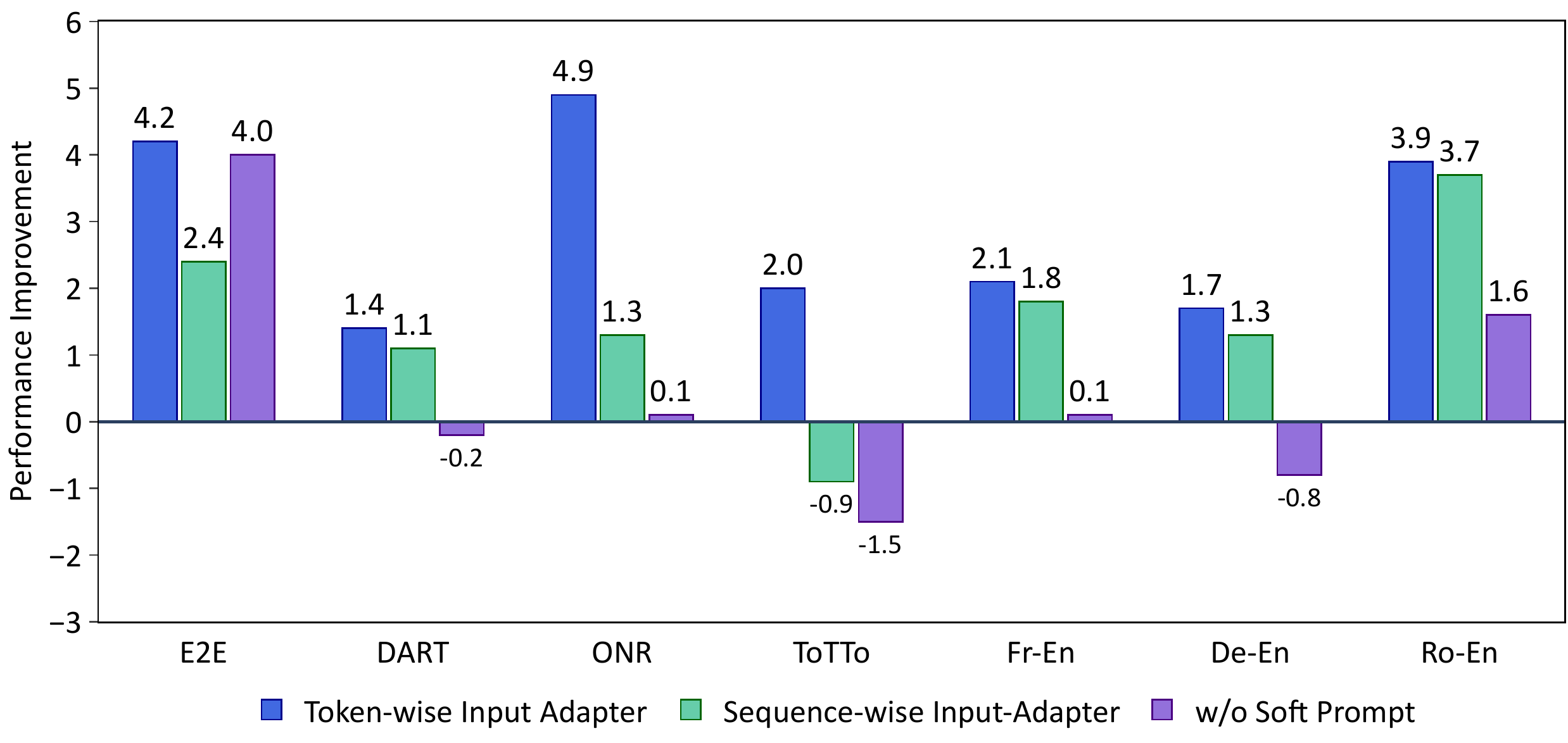}
    \caption{
    The improvements of input-tuning (w.r.t. prompt-tuning) with different designs.
    We take the PARENT score for ToTTo and BLEU scores for others.
    }
    \label{fig:ablation_all_tasks}
\end{figure*}

As shown in Figure~\ref{fig:ablation_all_tasks}, the sequence-wise input-adapter consistently performs worse the token-wise input-adapter, and without the soft prompts, the performance of input-tuning always drop.
These results enhance that a simple token-wise input-adapter can activate the frozen PLM better, and the soft prompts are indispensable in input-tuning.

\end{document}